%% file: main.tex
\documentclass[10pt,twocolumn,letterpaper]{article}

\usepackage{cvpr}              % To produce the CAMERA-READY version
\usepackage{cite}
\usepackage{amsmath,amssymb,amsfonts}
\usepackage{algorithm}
\usepackage{algorithmic}
\usepackage{booktabs}
\usepackage{graphicx}
\usepackage{textcomp}
\usepackage{xcolor}
\def\BibTeX{{\rm B\kern-.05em{\sc i\kern-.025em b}\kern-.08em
    T\kern-.1667em\lower.7ex\hbox{E}\kern-.125emX}}

\usepackage{pifont}
\newcommand{\cmark}{\textcolor{green!60!black}{\ding{51}}}
\newcommand{\xmark}{\textcolor{red}{\ding{55}}}
\usepackage{amsmath}
\usepackage{enumitem}
\usepackage{lineno}
\usepackage{tabularx}
\usepackage{makecell}
\usepackage{subcaption}
\usepackage{makecell}
\usepackage{fancyhdr}
\usepackage{amsfonts}
\usepackage{multirow}
\usepackage{comment}
\usepackage{subcaption}
\newcommand{\deemph}[1]{\textcolor{gray}{#1}}

\usepackage[pagebackref,breaklinks,colorlinks]{hyperref}

\usepackage[capitalize]{cleveref}
\crefname{section}{Sec.}{Secs.}
\Crefname{section}{Section}{Sections}
\Crefname{table}{Table}{Tables}
\crefname{table}{Tab.}{Tabs.}

\def\cvprPaperID{1842} % *** Enter the CVPR Paper ID here
\def\confName{WACV}
\def\confYear{2025}

\begin{document}

\title{Bridging Vision Language Models and Symbolic Grounding for Video Question Answering}

% \and
\author{
    Haodi Ma, Vyom Pathak, Daisy Zhe Wang \\
    Univerisy of Florida\\
    \texttt{\{ma.haodi, v.pathak, daisyw\}@ufl.edu}
}
\maketitle

%%%%%%%%% ABSTRACT
\begin{abstract}
    Video Question Answering (VQA) requires models to reason over spatial, temporal, and causal cues in videos. Recent vision language models (VLMs) achieve strong results but often rely on shallow correlations, leading to weak temporal grounding and limited interpretability. 

    We study symbolic scene graphs (SGs) as intermediate grounding signals for VQA. SGs provide structured object-relation representations that complement VLMs' holistic reasoning. We introduce SG-VLM, a modular framework that integrates frozen VLMs with scene graph grounding via prompting and visual localization. 

    Across three benchmarks (NExT-QA, iVQA, ActivityNet-QA) and multiple VLMs (QwenVL, InternVL), SG-VLM improves causal and temporal reasoning and outperforms prior baselines, though gains over strong VLMs are limited. These findings highlight both the promise and current limitations of symbolic grounding, and offer guidance for future hybrid VLM-symbolic approaches in video understanding.
\end{abstract}

%%%%%%%%% BODY TEXT
\section{Introduction}
\label{sec:intro}
\input{sections/1.intro}

\section{Related Works}
\label{sec:related_works}
\input{sections/2.related_works}

\section{Preliminary}
\label{sec:preliminary}
\input{sections/3.preliminary}

\section{Method}
\label{sec:method}
\input{sections/4.method}

\section{Experiment}
\label{sec:experiment}
\input{sections/5.experiments}

%%%%%%%%% REFERENCES
{\small
\bibliographystyle{ieee_fullname}
\bibliography{egbib}
}

\end{document}

%% file: sections/1.intro.tex
Video Question Answering (VQA) challenges models to understand and reason over complex visual and temporal content. While recent vision language models (VLMs)~\cite{li2023blip, dai2023instructblip, alayrac2022flamingo, Qwen2.5-VL, chen2024internvl, achiam2023gpt} achieve strong results in image and video-level reasoning, their performance often relies on shallow correlations rather than faithful multi-step reasoning. As recent studies~\cite{xiao2024nextgqa, pan2023retrieving, upadhyay2025time, rahmanzadehgervi2024vision} highlight, VLMs frequently suffer from hallucination and lack temporal or causal grounding, especially in long or complex videos.

One promising direction is to provide VLMs with intermediate grounding signals that decompose and clarify the reasoning process. Prior work has explored dense video captioning~\cite{min2024morevqa}, temporal span retrieval~\cite{yu2023self}, or caption-based retrieval~\cite{pan2023retrieving}. However, these approaches typically depend on textual descriptions or regional grounding, which may still lack structural transparency or fail to capture object-centric interactions essential for causal or spatial questions.

Scene graphs (SGs) have been widely studied in the context of images~\cite{johnson2015image, shi2019explainable}, where they provide structured object-relation representations that improve reasoning in image-based VQA, captioning, and retrieval. Scene graphs have also been proved to be valuable in commonsense and lightweight reasoning~\cite{wang2022sgeitl, nuthalapati2021lightweight}. Extending scene graph grounding from images to videos introduces new challenges: temporal dynamics, evolving object interactions, and cross-frame consistency. Several recent works attempt to construct video scene graphs for reasoning tasks, such as dynamic multi-step reasoning~\cite{mao2022dynamic} and graph-based temporal reasoning~\cite{urooj2023learning}, but these typically require training separate models or leveraging external detection, pre-existing SGs, and tracking pipelines. Such approaches are computationally expensive, less flexible, and harder to adapt across different VLM backbones.

In this work, we propose \textbf{SG-VLM}, a modular VQA framework that enhances frozen VLMs with \textit{symbolic scene graph grounding}. Unlike prior methods that depend on dedicated scene graph generators or additional training, SG-VLM directly leverages VLMs themselves to produce structured grounding through prompting. Our framework generates and selects question-relevant scene graphs over video frames to explicitly capture spatial and temporal object interactions that are essential for complex reasoning. The symbolic scene graphs serves as intermediate representations that provide interpretable grounding, and support multi-hop reasoning across temporal context.

Our framework consists of three stages: (1) Scene Graph Generation, where object-centric interactions are extracted using pre-trained VLMs; (2) Scene Graph Selection, which identifies question-relevant frames and associated graphs; and (3) Grounded Answer Generation, where video frames and symbolic groundings are combined for final prediction. We evaluate SG-VLM on three standard VQA benchmarks—NExT-QA~\cite{xiao2021next}, iVQA~\cite{yang2021just}, and ActivityNet-QA~\cite{yu2019activitynet}, covering open-ended, multiple-choice, and temporal reasoning tasks.

\textbf{In summary, this paper makes the following contributions:}
\begin{itemize}
    \item We conduct the first systematic evaluation of symbolic scene graphs under modern VLMs, benchmarking across three datasets and two strong backbones (QwenVL, InternVL).
    \item We formalize and evaluate four methods for integrating scene graphs into VQA: full SGs, question-based selection, in-range temporal extension, and SG summaries. This design space reveals trade-offs in coverage and reasoning capability.
    \item Our results show that scene graphs consistently improve causal and temporal reasoning and outperform prior baselines, but offer limited gains over strong VLMs. These findings highlight both the promise and current limitations of symbolic grounding, providing guidance for future hybrid VLM-symbolic approaches.
\end{itemize}

%% file: sections/2.related_works.tex
\subsection{Video Question Answering and Benchmarks}
Video Question Answering (VQA) tasks require models to comprehend visual content across time and answer natural language questions grounded in that content. Compared to static image VQA, video-based VQA introduces added challenges such as temporal reasoning, action tracking, and scene transitions. Several datasets have been proposed to benchmark progress. NExT-QA~\cite{xiao2021next} focuses on temporal and causal reasoning, requiring fine-grained comprehension of video events. ActivityNet-QA~\cite{yu2019activitynet} emphasizes question answering based on a large corpus of web videos spanning a broad range of human activities. iVQA~\cite{yang2021just} targets interactive video question answering, where questions are contextually grounded and often evolve with user interaction, testing both inference and generalization.

\subsection{End-to-End vision Language Models for VQA}
Recent progress in vision language pretraining has led to the development of end-to-end video question answering models that combine vision and language features using large-scale pretraining on image or video-text pairs. \textbf{Flamingo}\cite{alayrac2022flamingo} uses a frozen language model (LM) with learnable cross-modal layers, supporting few-shot visual reasoning over image and video inputs. \textbf{BLIP}\cite{li2022blip} introduces a query-aware cross-modal transformer between a frozen image encoder and a large LM, demonstrating strong performance on image-based VQA and captioning tasks. \textbf{Video-LLaMA}~\cite{zhang2023video} extends LLaMA for temporal understanding by fusing video frame representations into a frozen LM via projection and alignment layers. \textbf{CLIP}\cite{radford2021learning} and its video variants (e.g., VideoCLIP) learn cross-modal embeddings for retrieval-based or captioning-based QA.

Multimodal LLMs such as \textbf{GPT-4V}, \textbf{Qwen-VL} offer increasingly general-purpose capabilities for visual question answering, though their performance remains limited for tasks requiring structured, multi-hop, or temporal reasoning. These models often hallucinate visual content, as shown in recent studies~\cite{chen2024unified, hu2024improving, ullah2022thinking}, raising concerns about their factual consistency and visual grounding.

\subsection{Grounded and Adapted Reasoning Models}
To improve interpretability and reasoning accuracy and reduce hallucination, several works propose explicit grounding or model adaptation for VQA. \textbf{SeViLA}\cite{yu2023self} augments vision language models with grounding from video regions, improving explainability and localization. \textbf{NExT-GQA}\cite{xiao2024nextgqa} focuses on improving video understanding via visual grounding along the temporal dimension.

In parallel, other research explores adaptation strategies. \textbf{VisualGPT}\cite{chen2022visualgpt} integrates vision embeddings directly into GPT-2, enabling multi-modal response generation via finetuning. Language-based adaptation methods like \textbf{Retrieving-to-Answer}\cite{pan2023retrieving} use external video/text retrieval to prompt a frozen LLM with relevant captions, avoiding the need for full model finetuning. These approaches reduce training overhead and allow more flexible reasoning, but often depend heavily on caption quality and retrieval accuracy. Other works focus on symbolic grounding for images~\cite{nuthalapati2021lightweight, wang2022sgeitl} instead of videos, which doesn't capture essential cross-frame information in videos. In contrast, our approach grounds reasoning in symbolic structures, specifically, scene graphs extracted from selected video frames, rather than relying on bounding box grounding, extracted frames, or retrieved captions. 

%% file: sections/3.preliminary.tex
We consider the task of video question answering (VideoQA), where the input consists of a video $V = \{v_1, \dots, v_l\}$ containing $l$ frames, and a natural language question $Q$. The goal is to generate or select an answer $A$ that correctly responds to the question based on the video content. Depending on the setting, $A$ may either be an open-ended free-form response or a selected option from a predefined candidate set $A_{\text{cands}}$ ~\cite{xiao2024nextgqa, yang2021just}.

Formally, we define the VideoQA task as learning a function:
\begin{equation}
    M(V, Q, [A_{\text{cands}}], [V_{\text{groundings}}]) \rightarrow A,
\end{equation}
where $M$ is a multimodal reasoning model. The candidate set $A_{\text{cands}}$ is present for multiple-choice (closed-form) settings and omitted in open-ended formats. $V_{\text{groundings}}$ denotes any auxiliary visual grounding information—such as captions~\cite{pan2023retrieving} or event descriptions~\cite{min2024morevqa}—that may support intermediate reasoning or enhance model interpretability.

Our work builds upon this general formulation by introducing symbolic scene graph representations as intermediate visual grounding, enabling structured reasoning and interpretability.

%% file: sections/4.method.tex
\begin{figure*}
    \centering
    \includegraphics[width=\linewidth]{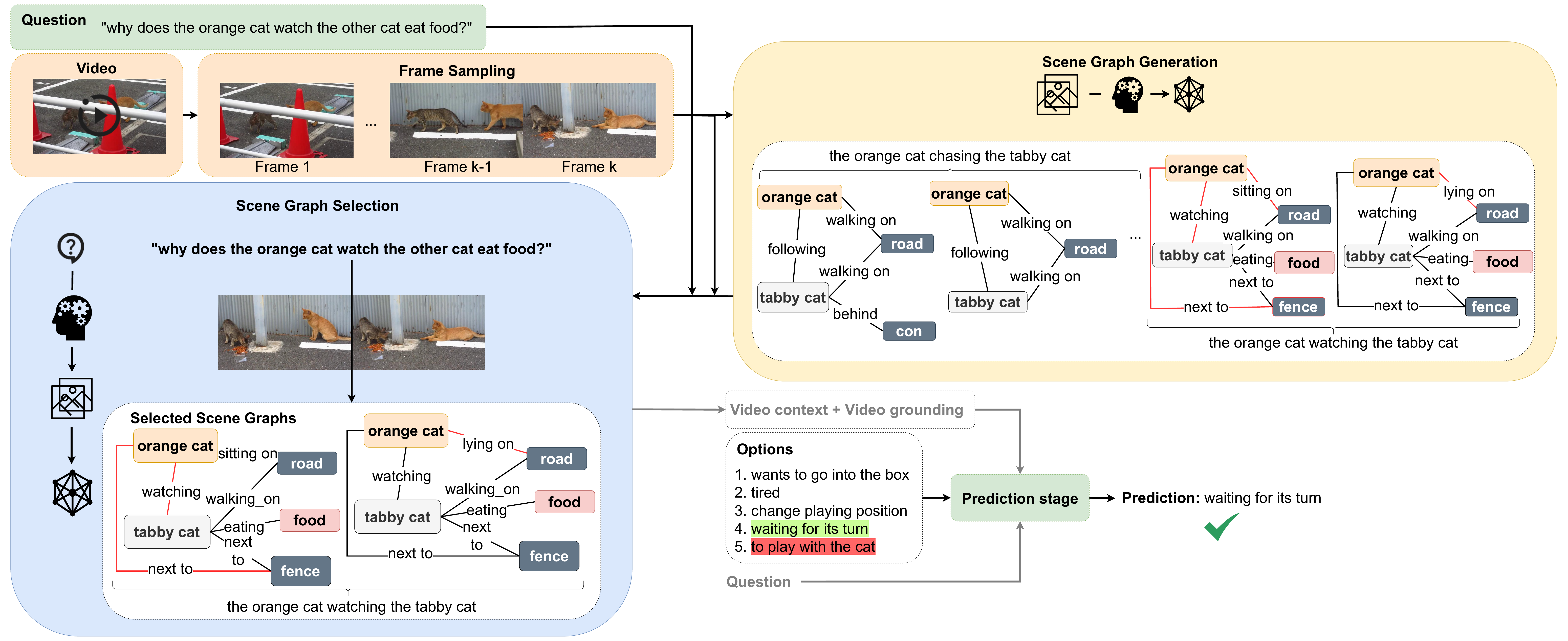}
    \caption{Overview of the SG-VLM pipeline with an illustrative example. Given the video and the question “why does the brown cat watch the other cat eat food?”, SG-VLM proceeds in three stages. (1) \textbf{Scene Graph Generation:} For each sampled frame, objects such as \emph{orange cat}, \emph{tabby cat}, and \emph{fence} are extracted, and spatial/action relations are constructed, e.g., (\emph{orange cat}, watching, \emph{tabby cat eating}). (2) \textbf{Scene Graph Selection:} Query-aware filtering retains only the graphs aligned with the question, discarding unrelated background relations. (3) \textbf{Grounded Answer Generation:} The selected graphs are combined with video context and used to prompt the VLM, leading to the correct prediction (\emph{“waiting for its turn”}). This process highlights how symbolic grounding complements VLM reasoning with interpretable object-relation structures.}
    \setlength{\belowcaptionskip}{-50pt}
    \label{fig:sgg-overview}
\end{figure*}

We present SG-VLM, a modular symbolic grounding framework designed to enhance vision language models (VLMs) in VQA. As illustrated in Figure~\ref{fig:sgg-overview}, SG-VLM introduces a structured intermediate reasoning layer via scene graphs to support more faithful, interpretable, and accurate answer generation. Unlike prior works that rely on separately trained scene graph models or external object detectors~\cite{mao2022dynamic}, our approach leverages frozen VLMs directly through prompting to construct symbolic grounding at each stage. This design makes SG-VLM lightweight, model-agnostic, and easily adaptable to different VLM backbones. The pipeline comprises three stages: (1) scene graph generation, (2) query-aware scene graph selection, and (3) grounded answer generation. To make the method concrete, we illustrate each stage using the example question: \emph{“why does the brown cat watch the other cat eat food?”}.

\subsection{Scene Graph Generation}
\label{sec:sgvlm-sgg}
Given an input video $V = \{v_1, \dots, v_l\}$, we sample representative frames $\{v_1, \dots, v_k\}$ where $k<l$ and construct per-frame scene graphs capturing objects and their interactions. Each graph consists of nodes (objects) and edges (relations), with two relation types: spatial and action-centric.

\subsubsection{Frame Sampling}
Frame sampling is crucial for balancing efficiency and coverage in long videos. Instead of evenly sampling frames from the given video, we explore a \textit{difference-based sampling} variant, which selects frames with the largest visual difference compared to their neighbors. This variant highlights dynamics, e.g., the moment when one cat stops walking and sits down, providing more informative symbolic groundings for temporally grounded questions.

\subsubsection{Object Identification}
We first prompt $m_{\text{VLM}}$ to produce structured descriptions for each frame and extract object mentions as candidate entities. 
Frequent objects across frames form the dominant set $O_{\text{main}}$, while co-occurring entities form $O_{\text{context}}$ in each frame. 
For example, in Figure~\ref{fig:sgg-overview}, the model identifies \emph{tabby cat} and \emph{orange cat} as $O_{\text{main}}$, while contextual entities such as \emph{road}, \emph{fence}, and \emph{food} appear as $O_{\text{context}}$ for each frame. This separation ensures that both central and supporting elements are represented in the scene graphs, enabling reasoning over interactions as well as background cues.

\subsubsection{Interaction Identification}
\paragraph{Spatial Relations}
We combine VLM-prompted object mentions with geometric cues to capture relative positioning. Bounding boxes are obtained via GroundingDINO~\cite{liu2024grounding} and refined with Segment Anything~\cite{ke2023segment}, then projected into 3D coordinates with off-the-shelf depth and camera models. Symbolic predicates such as \texttt{next to}, \texttt{behind}, or \texttt{above} are then assigned based on pairwise distances. 
In our example, the graphs include relations like (\emph{orange cat}, \texttt{next to}, \emph{fence}) and (\emph{tabby cat}, \texttt{on}, \emph{road}), explicitly encoding spatial layout.

\paragraph{Action-Centric Relations}
While spatial relations describe layout, action relations capture dynamics. We prompt $m_{\text{VLM}}$ with all detected objects $O$, overlaying bounding boxes to focus attention. Outputs are restricted to atomic triples \texttt{[subject, relation, object]}. For the cat example, this yields relations such as (\emph{orange cat}, \texttt{watching}, \emph{tabby cat}) and (\emph{tabby cat}, \texttt{eating}, \emph{food}), directly reflecting the causal setup in the question. These symbolic triples provide a compact, interpretable representation that complements continuous video features.

\subsubsection{Temporal Action Tracking}
Per-frame graphs capture local interactions but miss long-range dependencies. 
To enhance temporal grounding, we generate a global caption of the video to propose candidate actions, then verify their presence over time with a sliding window of size $k_2$. 
This produces a temporal action map—e.g., (\emph{orange cat}, \texttt{watching}, \emph{tabby cat}) persists across multiple frames—allowing the system to model both continuity and transitions of actions.

\subsection{Scene Graph Selection and Reasoning}
Not all scene graphs are equally useful for a given question. 
Directly feeding all graphs risks introducing noise and redundancy. 
To address this, we design a query-aware selection step. 
Given a question $Q$ and frames $\{v_1, \dots, v_k\}$, we prompt $m_{\text{VLM}}$ with $P$ to identify the most relevant frames, and retrieve their associated scene graphs as shown in~\ref{alg:frame_selection}. For the cat question, the system selects frames where the \emph{orange cat} is explicitly \texttt{watching} the \emph{tabby cat eating}, while discarding unrelated background relations such as (\emph{road}, \texttt{beside}, \emph{pole}). This step narrows the symbolic context to align closely with the semantics of the query.

\begin{algorithm}[H]
\caption{Frame selection and scene graph extraction}
\label{alg:frame_selection}
\begin{algorithmic}[1]
\STATE \textbf{Input:} Frame sequence $F$, question $Q$, processor, model, device
\STATE \textbf{Output:} Relevant frames $R$, scene graphs $G$
\STATE Initialize $R \gets [\,]$, $G \gets [\,]$
\FOR{each frame $f_i \in F$}
    \STATE $answer \gets$ \texttt{QwenGeneration}(\{$f_i$, ``Relevant to $Q$? Yes/No''\})
    \IF{$answer =$ ``Yes''}
        \STATE Append $f_i$ to $R$
        \STATE $graph \gets$ \texttt{QwenGeneration}(\{$f_i$, ``Extract scene graph for $Q$''\})
        \STATE Append $graph$ to $G$
    \ENDIF
\ENDFOR
\STATE \textbf{return} $R, G$
\end{algorithmic}
\end{algorithm}

\subsection{Grounded Answer Generation}
Finally, we integrate the selected scene graphs with the original video frames for answer generation. A prompt provides $m_{\text{VLM}}$ with question-aware scene graphs with the questions and options, if provided, to VLM for answer generation. For the running example, the model grounds the causal relation between the two cats and produces the correct answer: \emph{“waiting for its turn”}. This fusion demonstrates how symbolic grounding not only improves interpretability but also reinforces reasoning faithfulness, especially for causal and temporal questions.

%% file: sections/5.experiments.tex
\subsection{Datasets}
We evaluate our model on three widely used VideoQA benchmarks that cover diverse video domains, lengths, and reasoning types, making them well-suited for studying the role of symbolic grounding:

\subsubsection{NExT-QA~\cite{xiao2021next}}
NExT-QA is specifically designed to test temporal and causal reasoning. Each video clip (average length: 43 seconds) is paired with one question and five candidate answers. Following prior works~\cite{suris2023vipergpt}, we adopt the multiple-choice (MC) setting and report results on the 4,996 test video-question pairs. This benchmark is particularly relevant to SG-VLM, as many questions require modeling object interactions and event dependencies over time.

\subsubsection{iVQA~\cite{yang2021just}}
iVQA consists of 7-30 second video clips sampled from HowTo100M~\cite{miech2019howto100m}. Each video is paired with a question and a set of human-annotated ground-truth answers, along with five candidate options. We use the standard 2,001 testing examples. This dataset emphasizes compositional reasoning over human-object interactions and temporal sequences, which are challenging for symbolic grounding due to frequent fine-grained manipulations.

\subsubsection{ActivityNet-QA~\cite{yu2019activitynet}}
ActivityNet-QA provides 5,800 videos, each paired with 10 open-ended question-answer annotations. Videos average 180 seconds in length, and questions cover actions, objects, and temporal events. Unlike the other datasets, the task requires free-form answer generation without candidate options. We follow recent practice~\cite{pan2023retrieving, min2024morevqa} and report results using GPT-based answer similarity matching. This benchmark highlights the challenge of long-horizon reasoning, where scene graph selection and summarization are particularly important for efficiency.

\subsection{Baselines}
We compare SG-VLM against several strong baselines, grouped into three categories:

\subsubsection{VLM-only Baseline} 
To isolate the effect of symbolic grounding, we implement a baseline where a pretrained vision language model (VLM)~\cite{Qwen2.5-VL, chen2024internvl} is prompted directly with the video and question. This model does not utilize any scene graph information and serves as a reference point for evaluating the added value of our pipeline. For fair comparison, the same VLM backbone is used across all SG-VLM variants.

\subsubsection{End-to-End Video-Language Models (VLMs)}
These models represent the dominant paradigm in VideoQA: end-to-end architectures that directly encode video and text into a reasoning backbone. They serve as competitive state-of-the-art references:
\begin{itemize}
\item \textbf{BLIP-2}~\cite{li2022blip}: Bridges frozen image encoders with large language models through a query-aware cross-modal transformer. Originally designed for images, it can be adapted to video via frame sampling.
\item \textbf{Flamingo}~\cite{alayrac2022flamingo}: A few-shot capable VLM with cross-attention fusion layers for reasoning across sequences of frames.
\item \textbf{ViperGPT(+)}~\cite{suris2023vipergpt, min2024morevqa}: A modular reasoning framework that generates structured programs to solve visual questions. We include both the original ViperGPT and its multi-stage extension MoReVQA.
\end{itemize}

\subsubsection{Grounding-based or Retrieval-Augmented QA}
These methods are conceptually closest to SG-VLM, as they inject intermediate grounding into the reasoning pipeline:
\begin{itemize}
\item \textbf{Retrieving-to-Answer}~\cite{pan2023retrieving}: Selects relevant captions to prompt a frozen LLM, improving factual accuracy by anchoring reasoning to retrieved text.
\item \textbf{MoReVQA}~\cite{min2024morevqa}: Extends ViperGPT with multi-stage reasoning over extracted events and entity interactions, enhancing compositionality and interpretability.
\item \textbf{SeViLA}~\cite{yu2023self}: Integrates spatially grounded visual layouts into reasoning, aligning predicted answers with visual evidence.
\end{itemize}
Together, these baselines allow us to compare SG-VLM against (1) a controlled VLM-only setup, (2) state-of-the-art end-to-end systems, and (3) methods that share our motivation of augmenting VQA with intermediate grounding.

\subsection{Implementation Details}
We implement SG-VLM using Qwen2.5-VL~\cite{Qwen2.5-VL} and InternVL~\cite{chen2024internvl} as the unified vision language model for scene graph generation, symbolic reasoning, and final answer generation. All components are executed in a model-agnostic prompting pipeline without additional finetuning.

We sample $m=16$ frames per video by default, following the protocol in MoReVQA~\cite{min2024morevqa}. We explore a difference-based variant that selects frames with the largest visual difference from neighbors. For main object extraction, we set a frequency threshold of $p_1=0.6$ (objects must appear in at least 60\% of frames), and an object detection confidence threshold of $p_2=0.4$. Outputs are generated deterministically with decoding temperature set to 0.5. 

Bounding boxes are obtained with GroundingDINO~\cite{liu2024grounding}, refined with Segment Anything~\cite{ke2023segment}, and projected to 3D using Metric3Dv2, WildCamera~\cite{zhu2023tame}, and PerspectiveFields~\cite{jin2023perspective}. Each prompting stage (object identification, relation extraction, frame selection, and final answering) is implemented as an independent module. Beam size is set to 1 for decoding unless otherwise noted. All experiments are run on NVIDIA B200 GPUs (40GB). The average inference time for a full SG-VLM pipeline (16 frames) is approximately 30 seconds per video.

\paragraph{Scene Graph Variants}
To study the effect of symbolic grounding, we evaluate four SG integration strategies:
\begin{itemize}
\item \textbf{Full-SG}: all scene graphs from sampled frames are used.
\item \textbf{FrameSel-SG}: only scene graphs from question-relevant frames are used.
\item \textbf{RangeSel-SG}: question-relevant graphs plus a $m$-frame temporal window are included ($m$=3 by default).
\item \textbf{Summary-SG}: only unique objects are retained across frames, discarding relations.
\end{itemize}
These settings define the design space of symbolic grounding, enabling a systematic analysis of coverage, efficiency, and reasoning impact.

\textbf{Code and pretrained model calls will be made available upon publication.}

\begin{table*}[h]
\centering
\subcaptionbox{\centering NExT-QA~\cite{xiao2021next}}{
    \scalebox{0.71}{
        \begin{tabular}{l|cc}
            \toprule
             
            Method & Val  & FT \\
            \midrule
            \deemph{MIST-CLIP~\cite{gao2023mist}}
            & \deemph{57.2} &  \multirow{3}{*}{\cmark} \\
            \deemph{HiTeA~\cite{ye2023hitea}}         
            & \deemph{63.1}  &   \\
            \deemph{SeViLa~\cite{yu2023self}}
            & \deemph{73.8} &  \\
            \midrule
            \midrule
            ViperGPT~\cite{suris2023vipergpt}                
            & 60.0 & \multirow{9}{*}{\xmark} \\
            
            BLIP-2$^{\text{concat}}$~\cite{li2023blip}
            & 62.4 & \\
            
            BLIP-2$^{\text{voting}}$~\cite{li2023blip}
            & 62.7 &  \\
            SeViLA~\cite{yu2023self}
            & 63.6 &  \\\cline{1-2}\\[-1ex]
            
            FrameSel-SG + & & \\
            \hspace{3mm}Qwen2.5-VL-7B~\cite{Qwen2.5-VL}         & 77.8     & \\
            \hspace{3mm}Qwen2.5-VL-32B~\cite{Qwen2.5-VL}        & 76.8     & \\
            \hspace{3mm}InternVL-7B~\cite{chen2024internvl}     & \underline{81.1}     & \\
            \hspace{3mm}InternVL-14B~\cite{chen2024internvl}    & \textbf{83.6}     & \\
            \bottomrule
        \end{tabular}
    }
}
% \hfill
\subcaptionbox{\centering iVQA~\cite{yang2021just}}{
    \scalebox{0.71}{
        \begin{tabular}{l|cc}
            \toprule

            Method & Test & FT \\
            \midrule

            \deemph{VideoCoCa~\cite{yan2022videococa}} & \deemph{39.0}  & \multirow{3}{*}{\cmark} \\
            \deemph{FrozenBiLM~\cite{yang2022zero}} & \deemph{39.7}  &  \\
            \deemph{Text+Text}~\cite{lin2023towards} & \deemph{40.2}  &  \\

            \midrule
            FrozenBiLM~\cite{yang2022zero} & 27.3 & \multirow{9}{*}{\xmark} \\
            BLIP-2$_\text{\small{(FlanT5XXL)}}$~\cite{li2023blip}  & 45.8 &  \\
            InstructBLIP$_\text{\small{(FlanT5XL)}}$~\cite{dai2023instructblip}  & 53.1 &  \\
            InstructBLIP$_\text{\small{(FlanT5XXL)}}$~\cite{dai2023instructblip} & 53.8 &  \\\cline{1-2}\\[-1ex]
            FrameSel-SG + & & \\
            \hspace{3mm}Qwen2.5-VL-7B~\cite{Qwen2.5-VL}         & 68.6     & \\
            \hspace{3mm}Qwen2.5-VL-32B~\cite{Qwen2.5-VL}        & 70.2     & \\
            \hspace{3mm}InternVL-7B~\cite{chen2024internvl}     & \underline{73.1}     & \\
            \hspace{3mm}InternVL-14B~\cite{chen2024internvl}    & \textbf{76.9}     & \\
            \bottomrule
        \end{tabular}
    }
}
% \hfill
\subcaptionbox{\centering ActivityNet-QA~\cite{yu2019activitynet}}{
    \scalebox{0.71}{
        \begin{tabular}{l|cc}
            \toprule

            Method & Test & FT \\
            \midrule
            Video-LLaMA~\cite{zhang2023video} & 12.4 & \\
            VideoChat~\cite{li2023videochat} & 26.5 & \multirow{10}{*}{\xmark}  \\
            LLaMa adapter~\cite{zhang2023llamaadapter} & 34.2 &  \\ 
            Video-ChatGPT~\cite{maaz2023videochatgpt} & 35.2 &  \\  
            ViperGPT+ & 37.1 &  \\
            \cline{1-2}\\[-1ex]
            FrameSel-SG + & & \\
            \hspace{3mm}Qwen2.5-VL-7B~\cite{Qwen2.5-VL}         & 43.7  &   \\
            \hspace{3mm}Qwen2.5-VL-32B~\cite{Qwen2.5-VL}        & 43.9  &   \\
            \hspace{3mm}InternVL-7B~\cite{chen2024internvl}     & \underline{52.1}  &   \\
            \hspace{3mm}InternVL-14B~\cite{chen2024internvl}    & \textbf{52.7}  &   \\
            \bottomrule
            % \vspace{3mm}
        \end{tabular}
    }
}
% \hfill
% \vspace{-2.0mm}
\caption{\label{tab:sgvqa-sota_comparison}\textbf{Comparison to SOTA on the standard video question-answering datasets:} (a) NExT-QA, (b) iVQA, (c) ActivityNet-QA. FT indicates fine-tuned methods. 
}
\end{table*}

\subsection{Main Results}
\label{sec:agvqa-main_results}
Table~\ref{tab:sgvqa-sota_comparison} compares SG-VLM with FrameSel-SG against prior methods on three representative VideoQA benchmarks. 
Across all datasets, SG-VLM achieves strong performance and surpasses existing baselines, particularly when combined with larger VLM backbones.

NExT-QA emphasizes temporal and causal reasoning. FrameSel-SG substantially improves over classical modular reasoning systems such as ViperGPT (60.0\%) and SeViLA (63.6\%), reaching 83.6\% with InternVL-14B. The improvements are consistent across both Qwen and InternVL backbones, demonstrating that symbolic grounding complements pretrained VLMs for causal and temporal questions. 
Interestingly, Qwen-7B slightly outperforms Qwen-32B on NExT-QA, suggesting that symbolic grounding can sometimes reduce the performance gap between smaller and larger backbones. We hypothesize this may result from dataset-specific alignment, though the overall trend still favors larger models. The gap between 8B and 14B InternVL backbones remains visible, indicating that scaling the VLM is still a major factor.

On iVQA, which focuses on human-object interactions and temporal sequences, SG-VLM also provides significant gains. FrameSel-SG achieves up to 76.9\% (InternVL-14B), outperforming InstructBLIP (53.8\%) and other caption-based systems by a large margin. This suggests that scene graph grounding offers valuable structure in settings where fine-grained human-object interactions must be tracked.

For ActivityNet-QA with long videos and open-ended questions, SG-VLM reaches 52.7\% with InternVL-14B, outperforming Video-ChatGPT (35.2\%) and ViperGPT+ (37.1\%). The improvements highlight the role of scene graph selection in filtering noise from long contexts, making symbolic grounding particularly useful for efficiency in long-horizon reasoning.

Across all three benchmarks, FrameSel-SG consistently outperforms prior baselines and achieves strong results across both QwenVL and InternVL families. While performance generally scales with model size (e.g., InternVL-14B surpassing InternVL-7B), we also observe that Qwen-7B slightly outperforms Qwen-32B on NExT-QA, suggesting that symbolic grounding can help smaller backbones close the gap in certain settings. Importantly, SG-VLM outperforms previous methods regardless of model scale, confirming that symbolic scene graphs provide complementary benefits to pretrained VLMs in causal, temporal, and long-horizon reasoning tasks.

\begin{table}[h]
\large
\centering
\caption{Performance comparison of different settings of SG-VLM with 2 backbone VLMs.}
\label{table:quality_evaluation}
    \resizebox{\columnwidth}{!}{
        \begin{tabular}{ll|c|c|c} 
        \toprule
        VLM & Setup & NExT-QA & iVQA & ActivityNet-QA \\
        \midrule
        \multirow{5}{*}{Qwen2.5VL-7B}   & No SG                 & 79.5  & 69.1  & 46.6  \\
                                        & Full SG               & 74.1  & 68.5  & 38.8  \\
                                        & FrameSel-SG           & 77.8  & 68.6  & 43.7  \\
                                        & RangeSel-SG           & 74.6  & 68.2  & 41.6  \\
                                        & Summary-SG            & 77.9  & 67.3  & 44.7  \\
        \midrule
        \multirow{5}{*}{Qwen2.5VL-32B}  & No SG                 & 78.4  & 69.6  & 44.7  \\
                                        & Full SG               & 75.9  & 69.6  & 43.1  \\
                                        & FrameSel-SG           & 76.8  & 70.2  & 43.9  \\
                                        & RangeSel-SG           & 74.7  & 70.6  & 41.8  \\
                                        & Summary-SG            & 77.7  & 71.7  & 44.8  \\
        \midrule
        \multirow{5}{*}{InternVL-8B}    & No SG                 & 84.1  & 77.4  & 54.6  \\
                                        & Full SG               & 80.6  & 73.5  & 52.0  \\
                                        & FrameSel-SG           & 81.1  & 73.1  & 52.1  \\
                                        & RangeSel-SG           & 74.3  & 67.0  & 46.5  \\
                                        & Summary-SG            & 83.1  & 70.8  & 51.4  \\
        \midrule
        \multirow{5}{*}{InternVL-14B}   & No SG                 & 86.0  & 77.5  & 54.6  \\
                                        & Full SG               & 83.1  & 76.8  & 52.9  \\
                                        & FrameSel-SG           & 83.6  & 76.9  & 52.7  \\
                                        & RangeSel-SG           & 77.3  & 71.3  & 48.3  \\
                                        & Summary-SG            & 85.1  & 75.0  & 52.9  \\
        \bottomrule
    \end{tabular}
}
\end{table}

\subsection{Ablation on SG Variants}
Table~\ref{table:quality_evaluation} reports results across four SG integration strategies compared to VLM-alone (No SG) across all 3 datasets. We find that the four SG settings reveal important trade-offs:
\begin{itemize}
    \item \textbf{Selection vs. Full SG.} Across all datasets and backbones, FrameSel-SG consistently outperforms Full-SG. This demonstrates the necessity of question-aware localization: using all graphs introduces noise from irrelevant frames, while targeted selection retains only the most useful symbolic context.
    \item \textbf{Summary-SG.} In many cases, Summary-SG is competitive with or better than Full-SG, and sometimes even stronger than FrameSel-SG (e.g., Qwen-32B on iVQA). This suggests that object mentions are often more reliably extracted than fine-grained relations, so removing noisy edges can reduce error propagation.
    \item \textbf{RangeSel-SG.} Extending selection with neighboring frames generally hurts performance, indicating that the added temporal context often introduces spurious objects or actions. This highlights that SG quality, rather than quantity, is the key bottleneck.
\end{itemize}
Overall, these comparisons show that our system design choices matter: selection and summarization mitigate some noise, but further improvement in SG extraction is critical.

The effect of symbolic grounding varies across datasets. On NExT-QA, symbolic variants underperform the strong VLM-only baselines. Since the clips are moderately long but still well-structured, pretrained VLMs already capture much of the spatio-temporal context, and the additional symbolic graphs sometimes conflict with these internal priors, leading to accuracy drops. A similar trend is observed on ActivityNet-QA, where long videos (average 180 seconds) make SG extraction more error-prone. Here, symbolic graphs often fail to capture the long-horizon dependencies needed for open-ended questions, and noisy object-relation triples may dilute the VLM’s reasoning ability. In contrast, iVQA shows a different pattern: for Qwen2.5VL-32B backbones, all SG variants outperform the VLM-only baseline, with FrameSel-SG and Summary-SG yielding the strongest results. Instructional, step-by-step videos benefit more from explicit object and interaction grounding, which helps filter distractions and anchor reasoning around the relevant entities. And as the videos are shorter, the dynamics are better captured within the symbolic representations, yielding stronger performance.

These findings suggest that scene graphs are not universally beneficial when applied in a plug-and-play manner, but the comparison among variants reveals several important insights. First, question-aware selection is essential: filtering irrelevant frames consistently outperforms using all graphs. Second, object mentions tend to be more reliably extracted than fine-grained actions and relations, so summarization sometimes improves robustness by reducing noise from imperfect relations. Third, symbolic context proves most useful in domains requiring step-by-step human-object reasoning, as seen in iVQA. Taken together, these results indicate that while symbolic grounding is currently limited by SG quality, it provides complementary benefits and interpretability, and future work should explore more robust relation extraction and adaptive temporal modeling to further unlock its potential.

\subsection{Ablation on Question-Type Analysis}
\label{sec:sgvqa-ablation_analysis}
To better understand the role of symbolic grounding, we analyze performance by question type on NExT-QA (Table~\ref{tab:sgvqa-ablation-nextqa-question_type}). Although the VLM-only baseline achieves the highest overall accuracy (78.4\%), different SG variants show complementary strengths across specific categories, revealing both the promise and the current limitations of symbolic grounding.

The Summary-SG setting, which retains only object mentions and discards relations, outperforms VLM-only on Descriptive Open (+2.3) and Temporal Current (+2.9). This indicates that object detection is relatively robust, and simplifying the graph to objects alone reduces noise from imperfect or spurious relations. 
These gains are most evident in object or state centric questions (e.g., “What is the man holding?”), where recognizing the correct entities is more important than modeling detailed interactions. The improvements suggest that object-only grounding can serve as a reliable symbolic clue to better ground visual details for VLMs.

FrameSel-SG, which selects scene graphs from question-relevant frames, achieves the best results on Descriptive Count (+2.2) and Temporal Next (+3.3). This demonstrates that visual localization is crucial: irrelevant frames dilute reasoning with distracting objects and relations, while focusing on relevant slices enhances precision. Counting questions benefit from filtering, since duplications or extraneous objects are excluded. Similarly, “what happens next” questions require temporal specificity, and localizing the graph helps VLMs focus on the right part of the video. The consistent superiority of FrameSel-SG over Full-SG and even VLM themselves further highlights that selection is necessary for effective symbolic grounding.

Although noisier overall, Action-SG, with only actions in each frame provided, shows potential for temporal reasoning categories such as Temporal Next and Temporal Previous. Explicit actions sequence (e.g., "The orange cat following the tabby cat" to “The orange cat watching the tabby cat”) provide interpretable signals about event ordering, which can complement the implicit sequence modeling of VLMs. However, inaccuracies in relation extraction limit its effectiveness. As a result, Action-SG lags behind in descriptive and causal categories but points toward the value of more robust action-centric grounding for temporal tasks.

At the same time, symbolic variants underperform on the largest category, Causal Why, where VLMs already achieve strong results. Because these questions dominate the dataset (over 1900 examples), small drops here outweigh improvements in less frequent categories. This reflects a broader challenge: when VLMs are already strong on certain reasoning types, additional symbolic input can introduce redundancy or noise, lowering the overall score. Nonetheless, the category-level analysis highlights that symbolic grounding provides complementary benefits: object-only grounding helps descriptive and current questions, frame selection improves counting and next-step reasoning, and relations offer promise for temporal ordering. Improving the quality of relation extraction and better handling of causal questions remain key directions for making symbolic graphs consistently beneficial.

\begin{table*}[h]
\centering
    \begin{tabular}{c|ccccccccc}
        \toprule         
          & CH  & CW & DC & DL & DO & TC & TN & TP & Total  \\
         \midrule
         Question Count     & 683  & 1924 & 177 & 295 & 305 & 663 & 895 & 54 & 4996  \\
         \midrule
         \midrule
         No SG              & \textbf{79.2}  & \textbf{78.6} & 67.8 & \textbf{93.5} & 83.9 & 77.2 & 68.2 & \textbf{77.7} & \textbf{78.4} \\
         Full SG            & 78.6  & 76.2 & 63.8 & 89.8  & 83.0 & 76.2 & 68.6 & 72.2 & 75.9\\
         FrameSel SG        & 76.6  & 77.8 & \textbf{70.0} & 90.9 & 84.6 & 79.6 & \textbf{71.5} & 68.5 & 76.8 \\
         Summary SG         & 77.8  & 78.2 & 68.4 & 91.5  & \textbf{86.2} & \textbf{80.1} & 68.6 & 72.2 & 77.7 \\
         Action SG          & 78.2  & 76.3 & 66.8 & 89.8  & 83.0 & 76.2 & 70.6 & 72.2 & 77.1 \\
         \bottomrule
    \end{tabular}
\caption{Performance for each type of quesitons on NExt-QA: Causal How (CH), Causal Why (CW), Desc. Count (DC), Desc. Location (DL), Desc. Open (DO), Temporal Current (TC), Temporal Next (TN), Temporal Previous (TP).}
\label{tab:sgvqa-ablation-nextqa-question_type}
\end{table*}

\section{Conclusion}
In this work, we presented SG-VLM, a modular framework that integrates symbolic scene graphs into frozen vision language models for video question answering. Through comprehensive experiments on three benchmarks and multiple VLM backbones, we provided the first systematic study of how symbolic grounding interacts with VLMs. Our results show that while scene graphs do not consistently outperform strong VLMs overall, they provide clear benefits for specific reasoning categories: object-only graphs improve descriptive and current questions, frame selection enhances counting and next-step prediction, and relation-based graphs show potential for temporal ordering. We also found that question-aware selection is essential, and that noisy relation extraction remains a bottleneck, particularly for causal questions.

These findings highlight both the promise and the limitations of symbolic grounding in the VLM era. 
Scene graphs remain valuable for interpretability and targeted reasoning, and our analysis provides guidance on when and how they should be applied. Future work should focus on improving the quality of relation and action extraction, exploring adaptive integration strategies that dynamically decide when symbolic grounding is beneficial, and extending symbolic methods to capture causal and long-horizon dependencies more faithfully.